\documentclass[letterpaper, 10 pt, conference]{ieeeconf}
\IEEEoverridecommandlockouts                              
\usepackage{times}
\usepackage{epsfig}
\usepackage{graphicx}
\usepackage{multirow}
\usepackage{multicol}

\usepackage{comment}
\usepackage{url}

\usepackage{graphicx}
\usepackage{epsfig}
\usepackage{epic,eepic}
\usepackage{times}
\usepackage{mathptmx}
\usepackage{cite}
\usepackage{color}

\usepackage{times}

\definecolor{red}{rgb}{1,0,0}
\definecolor{green}{rgb}{0,1,0}
\definecolor{blue}{rgb}{0,0,1}
\definecolor{violet}{rgb}{1,0,1}
\definecolor{cyan}{cmyk}{1,0,0,0}
\definecolor{magenta}{cmyk}{0,1,0,0}
\definecolor{yellow}{cmyk}{0,0,1,0}

\definecolor{white}{rgb}{1,1,1}

\usepackage{arydshln}

\newcommand{\CO}[1]{}

\newcommand{\CommentOut}[1]{}

 \newcommand{\editage}[1]{}

\begin{document}

\title{\LARGE \bf
A Multi-modal Approach to Single-modal Visual Place Classification
}

\author{%
Tomoya Iwasaki$^{*}$, Kanji Tanaka$^{*}$, and Kenta Tsukahara$^{*}$
\thanks{$*$T. Iwasaki, K. Tanaka, and K. Tsukahara are with Department of Engineering, University of Fukui, Japan. {\tt\small tnkknj@u-fukui.ac.jp}}}

\maketitle


\newcommand{\TAB}[1]{#1}

\newcommand{\FIG}[3]{
\begin{minipage}[b]{#1cm}
\begin{center}
\includegraphics[width=#1cm]{#2}\\
{\scriptsize #3}
\end{center}
\end{minipage}
}

\newcommand{\FIGU}[3]{
\begin{minipage}[b]{#1cm}
\begin{center}
\includegraphics[width=#1cm,angle=180]{#2}\\
{\scriptsize #3}
\end{center}
\end{minipage}
}

\newcommand{\FIGm}[3]{
\begin{minipage}[b]{#1cm}
\begin{center}
\includegraphics[width=#1cm]{#2}\\
{\scriptsize #3}
\end{center}
\end{minipage}
}

\newcommand{\FIGR}[3]{
\begin{minipage}[b]{#1cm}
\begin{center}
\includegraphics[angle=-90,width=#1cm]{#2}
\\
{\scriptsize #3}
\vspace*{1mm}
\end{center}
\end{minipage}
}

\newcommand{\FIGRpng}[5]{
\begin{minipage}[b]{#1cm}
\begin{center}
\includegraphics[bb=0 0 #4 #5, angle=-90,clip,width=#1cm]{#2}\vspace*{1mm}
\\
{\scriptsize #3}
\vspace*{1mm}
\end{center}
\end{minipage}
}

\newcommand{\FIGCpng}[5]{
\begin{minipage}[b]{#1cm}
\begin{center}
\includegraphics[bb=0 0 #4 #5, angle=90,clip,width=#1cm]{#2}\vspace*{1mm}
\\
{\scriptsize #3}
\vspace*{1mm}
\end{center}
\end{minipage}
}

\newcommand{\FIGpng}[5]{
\begin{minipage}[b]{#1cm}
\begin{center}
\includegraphics[bb=0 0 #4 #5, clip, width=#1cm]{#2}\vspace*{-1mm}\\
{\scriptsize #3}
\vspace*{1mm}
\end{center}
\end{minipage}
}

\newcommand{\FIGtpng}[5]{
\begin{minipage}[t]{#1cm}
\begin{center}
\includegraphics[bb=0 0 #4 #5, clip,width=#1cm]{#2}\vspace*{1mm}
\\
{\scriptsize #3}
\vspace*{1mm}
\end{center}
\end{minipage}
}

\newcommand{\FIGRt}[3]{
\begin{minipage}[t]{#1cm}
\begin{center}
\includegraphics[angle=-90,clip,width=#1cm]{#2}\vspace*{1mm}
\\
{\scriptsize #3}
\vspace*{1mm}
\end{center}
\end{minipage}
}

\newcommand{\FIGRm}[3]{
\begin{minipage}[b]{#1cm}
\begin{center}
\includegraphics[angle=-90,clip,width=#1cm]{#2}\vspace*{0mm}
\\
{\scriptsize #3}
\vspace*{1mm}
\end{center}
\end{minipage}
}

\newcommand{\FIGC}[5]{
\begin{minipage}[b]{#1cm}
\begin{center}
\includegraphics[width=#2cm,height=#3cm]{#4}~$\Longrightarrow$\vspace*{0mm}
\\
{\scriptsize #5}
\vspace*{8mm}
\end{center}
\end{minipage}
}

\newcommand{\FIGf}[3]{
\begin{minipage}[b]{#1cm}
\begin{center}
\fbox{\includegraphics[width=#1cm]{#2}}\vspace*{0.5mm}\\
{\scriptsize #3}
\end{center}
\end{minipage}
}







\newcommand{\figA}{
\begin{figure}
\begin{center}
~~~~\FIG{8}{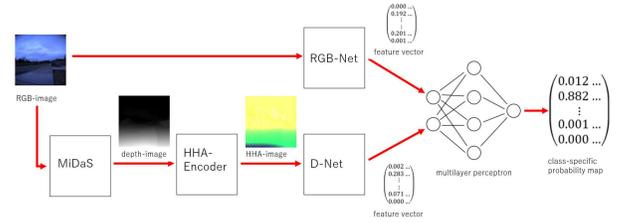}{}
\caption{Pseudo RGB-D multimodal framework.}\label{fig:A}
\end{center}
\end{figure}
}

\newcommand{\figB}{
\begin{figure}
\begin{center}
\hspace*{1cm}\FIG{8}{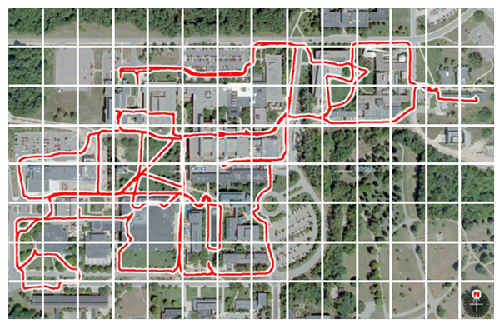}{}\vspace*{1cm}\\
\caption{A top-down view of the robot workspace with a grid of predefined place classes.}\label{fig:B}
\end{center}
\end{figure}
}

\newcommand{\figC}{
\begin{figure}
\begin{center}
~~~~\FIG{8}{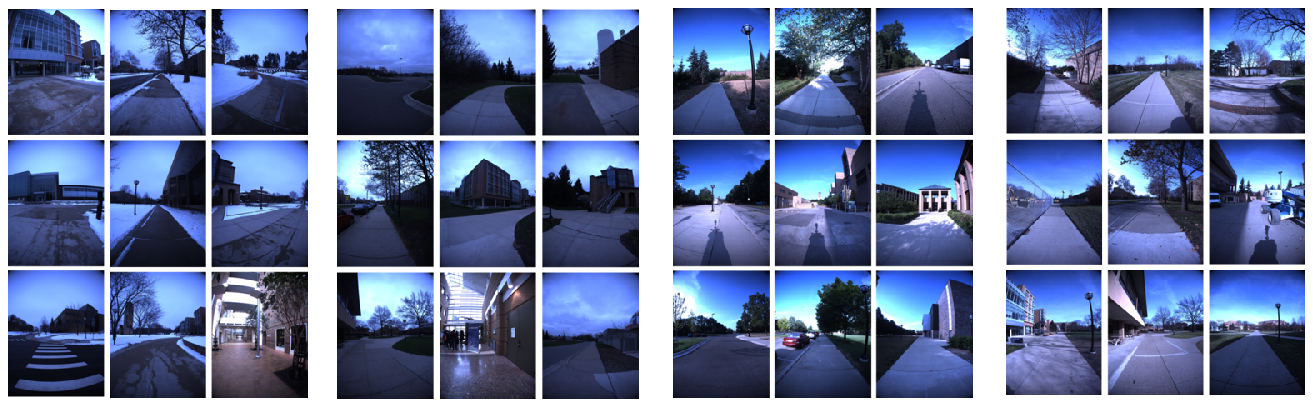}{}
\caption{Image samples from datasets ``WI," ``SP," ``SU," and ``AU".}\label{fig:C}
\end{center}
\end{figure}
}

\newcommand{\figD}{
\begin{figure}
\begin{center}
\hspace*{1cm}\FIG{8}{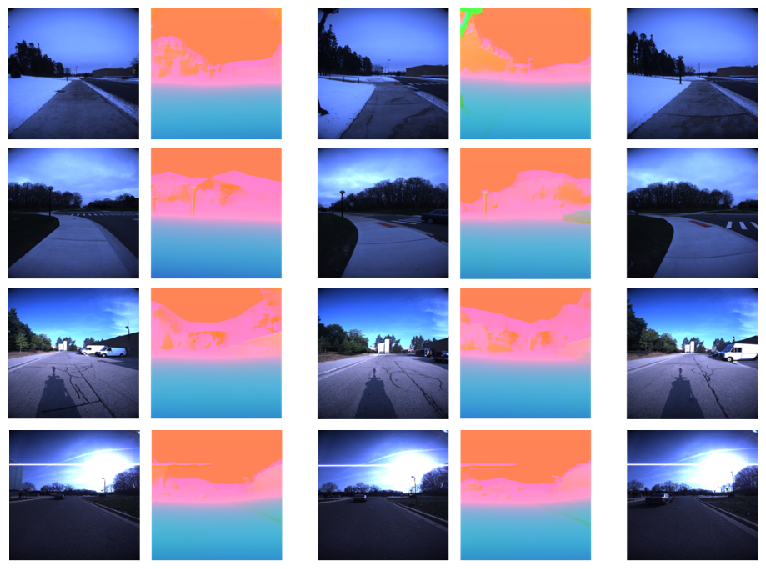}{}
\caption{Success examples.}\label{fig:D}
\end{center}
\end{figure}
}

\newcommand{\figE}{
\begin{figure}
\begin{center}
\hspace*{1cm}\FIG{8}{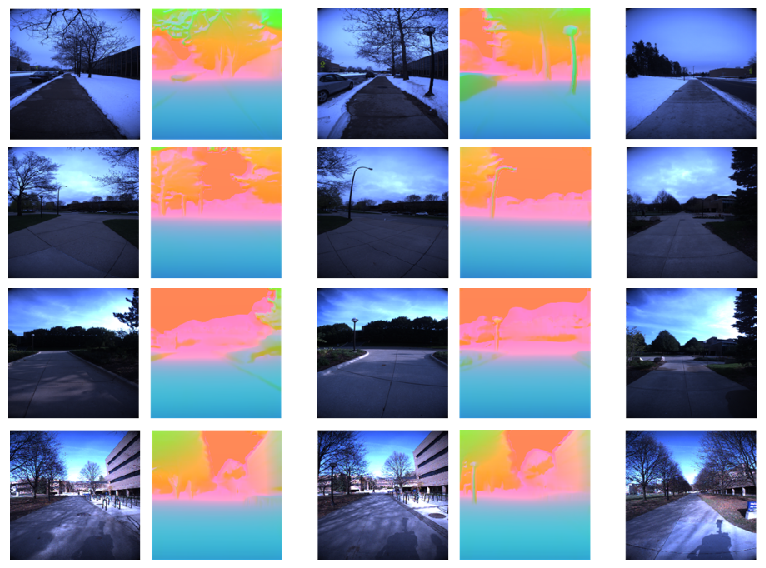}{}
\caption{Failure examples.}\label{fig:E}
\end{center}
\end{figure}
}

\newcommand{\figF}{
\begin{figure}
\begin{center}
\hspace*{1cm}\FIG{8}{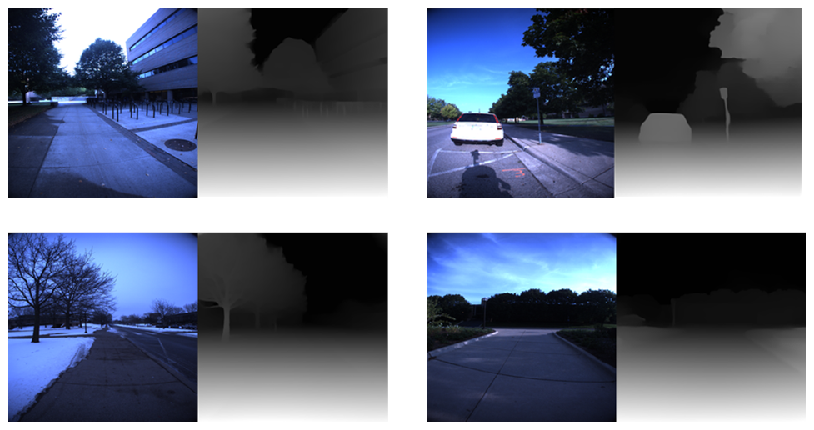}{}
\caption{Pseudo depth images.}\label{fig:F}
\end{center}
\end{figure}
}

\newcommand{\tabA}{
\begin{table*}
\begin{center}
\caption{Top-1 accuracy [\%].}\label{tab:A}
\TAB{
\begin{tabular}{|l|l|r|rr|rr|}
\hline
training & test~~~~ & 
~~~~Ours & ~~~~RGB-Net & (gain) & ~~~~HHA-Net & (gain) \\
\hline
\multirow{3}{3em}{1/22} & 3/31 & 62.4 & 58.5 & +3.9 & 56.3 & +6.1 \\
& 8/4 & 49.1 & 40.2 & +8.9 & 43.4 & +5.7 \\
& 11/17 & 40.6 & 31.7 & +8.9 & 37.8 & +2.8 \\
\hline
\multirow{3}{3em}{3/31} & 1/22 & 60.4 & 48.7 & +11.7 & 55.3 & +5.1 \\
& 8/4 & 59.3 & 47.1 & +12.2 & 52.9 & +6.4 \\
& 11/17 & 40.6 & 27.1 & +13.5 & 38.3 & +2.3 \\
\hline
\multirow{3}{3em}{8/4} & 1/22 & 42.4 & 32.6 & +9.8 & 40.0 & +2.4 \\
 & 3/31 & 57.8 & 49.2 & +8.6 & 49.9 & +7.9 \\
 & 11/17 & 37.2 & 26.3 & +10.9 & 31.3 & +5.9 \\
\hline
\multirow{3}{3em}{11/17}  & 1/22 & 41.3 & 29.3 & +12 & 39.1 & +2.2 \\
 & 3/31 & 48.8 & 38.2 & +10.6 & 41.5 & +7.3 \\
 & 8/4 & 38.8 & 29.2 & +9.6 & 32.1 & +6.7 \\
\hline
\end{tabular}
}
\end{center}
\end{table*}
}

\begin{abstract}
Visual place classification from a first-person-view monocular RGB image is a fundamental problem in long-term robot navigation. A difficulty arises from the fact that RGB image classifiers are often vulnerable to spatial and appearance changes and degrade due to domain shifts, such as seasonal, weather, and lighting differences. To address this issue, multi-sensor fusion approaches combining RGB and depth (D) (e.g., LIDAR, radar, stereo) have gained popularity in recent years. Inspired by these efforts in multimodal RGB-D fusion, we explore the use of pseudo-depth measurements from recently-developed techniques of ``domain invariant" monocular depth estimation as an additional pseudo depth modality, by reformulating the single-modal RGB image classification task as a pseudo multi-modal RGB-D classification problem. Specifically, a practical, fully self-supervised framework for training, appropriately processing, fusing, and classifying these two modalities, RGB and pseudo-D, is described. Experiments on challenging cross-domain scenarios using public NCLT datasets validate effectiveness of the proposed framework.
\end{abstract}


\section{
Introduction
}

Self-localization from a first-person-view monocular RGB image is a fundamental problem in visual robot navigation, with important applications such as first-person-view 
point-goal navigation \cite{pgn},
vision-language navigation \cite{vln},
and
object-goal navigation \cite{ogn},
which has recently emerged in the robotics and vision communities. 
It is typically formulated as a task of visual place classification \cite{planet}, 
where the goal is to classify a first-person-view image into one of predefined place classes. This is a problem domain to which supervised or self-supervised learning is directly applicable and has become a predominant approach \cite{icra2019sc}.

A difficulty arises from the fact that a self-localization model is often trained and tested in different domains. Domain shifts due to such as seasonal, weather, and lighting differences often degrade a self-localization model that is overfitted to the training domain and that is sensitive to viewpoint and appearance changes. Hence, domain-invariant and domain-adaptive models are desirable. In machine learning, this is most relevant to an open issue, called ``domain adaptation" \cite{dasurvey}, which aims to address the shortage of large amounts of labeled data, by using various types of transfer learning techniques, ranging from feature distribution alignment to model pipeline modification.

This work is inspired by recent research efforts to solve this problem using RGB and depth (D) sensor fusion. The key idea is to combine the RGB image modality with other depth sensors (e.g., LIDAR, radar, stereo), to address the ill-posed-ness of monocular vision. It has become clear that these additional depth measurements provide effective invariant cues for self-localization, such as viewpoint invariant 3D structure of landmark objects. As a downside, these methods rely on additional sensing devices, limiting their versatility and cost. Nevertheless, the domain invariance of depth measurements makes them very attractive for cross-domain self-localization. 

\figA

Based on the consideration, we revisit the long-term 
single-modal RGB visual place classification from a novel perspective of multi-modal RGB-D sensor fusion (Fig. \ref{fig:A}). 
Instead of requiring additional sensing devices as most existing multi-sensor fusion schemes do, 
we 
propose 
to transform
the available RGB image 
to a pseudo depth (D) image (e.g., Fig. \ref{fig:F})
and then
reformulate the single-modal image classification task
as a pseudo multimodal classification problem.
Specifically, in our approach,
a CNN -based place classifier is trained for 
each of these two modalities, RGB and D, and then the two CNNs are integrated by a multi-layer perceptron. The two CNNs 
could be supervised, diagnosed and retrained independently, which allows flexible and versatile design for domain adaptation scheme. Experiments on challenging cross-domain self-localization scenarios using public NCLT (University of Michigan North Campus Long-Term Vision and Lidar) dataset \cite{nclt} 
validate effectiveness of the proposed framework.

The contributions of this research are summarized below.
(1)
We address the underexplored ill-posed-ness of the long-term single-modal visual place classification problem from a novel perspective of multimodal RGB-D fusion.
(2)
We present a novel multimodal CNN architecture for RGB-D fusion using pseudo D images from domain-invariant monocular depth estimation.
(3)
Experiments using the public NCLT dataset show that the proposed method 
frequently contributes to performance improvement.

This paper is organized as follows. 
Section \ref{sec:related} 
gives a short overview of related works. 
Section \ref{sec:problem}
formulates 
the problem of
single-modal visual place classification
in the context of long-term robot navigation.
Section \ref{sec:approach}
describes the proposed framework 
for multi-modal extension of the single modal classifier.
Section \ref{sec:exp}
presents and discusses the experimental results.
Finally, 
concluding remarks
are given in Section \ref{sec:conclusions}.

\section{
Related Work
}\label{sec:related}

The problem of self-localization has been extensively researched in numerous indoor and outdoor applications with various formulations such as image retrieval \cite{fabmap}, %
geometric matching \cite{vsl}, 
loop closure detection \cite{ibowlcd}, %
place classification \cite{planet}, and viewpoint regression \cite{regress}. 
This work focuses on the classification formulation, where the goal is to classify a first-person-view image into one of predefined place classes. This is a problem domain where supervised or self-supervised learning is directly applicable and has become a predominant approach \cite{icra2019sc}. 

Multimodal RGB-D sensor fusion is one of the most active research areas of cross-domain self-localization. In \cite{lidarradar}, 
lidar and radar were thoroughly compared in terms of cross-season self-localization performance. In \cite{rgbs}, a highly robust scheme for long-term self-localization was explored where a semantic-geometric model reconstructed from RGB-D and semantic (RGB-D-S) images with a prior map. In \cite{lidarad}, 
a highly versatile self-localization framework for autonomous driving with LIDAR sensors was constructed. 
In \cite{ltlcd}, 
simultaneous training and deployment of an online self-localization task called loop closure detection was explored using LIDAR and imagery in a long-term map maintenance scenario. It is clear that RGB-D fusion is effective for achieving a good trade-off between robustness and accuracy in cross-domain scenarios.

In existing studies of cross-domain multi-modal (``RGB-X") visual place classification, so far, monocular depth estimation has not been fully explored as an additional modality (``X"). The main reason is that the technology for monocular depth estimation with domain invariance has not been established, until recently. Furthermore, many existing studies on cross-domain self-localization 
belong to image retrieval and matching paradigms, 
rather than the classification paradigm,
which was enabled by 
the recent advance of deep learning technology.

It should be noted
that
not depth,
but also 
other types of additional modalities are gaining in popularity.
Especially,
in the era of deep learning, 
semantic imagery
from deep semantic models
is 
one of such recently popular modalities.
In parallel with this work, we are also conducting research in that direction \cite{icte2023ohta}. However, many studies rely on depth measurements derived from prior or 3D reconstructions such as 3D point cloud maps, which is not assumed in this work. 
In addition, the semantic feature approach and our pseudo-depth approach are orthogonal and complementary.

Finally,
monocular depth estimation has received a great deal of attention in recent years in the machine learning and computer vision communities. Early work on monocular depth estimation 
used 
simple geometric assumptions,
non-parametric methods,
or
MRF-based formulations
 \cite{midas3}. 
More recently, 
the advent of powerful convolutional networks 
has enabled 
to directly regress D images
from RGB images \cite{midas15}. 
However, most existing
methods assume the availability of additional sensor modalities in the training stage as a means of self-supervised domain adaptation \cite{midas9}, 
which is not available in our case.
Our work is most relevant to and developed upon the recently developed technique of ``domain-invariant" monocular depth estimation in \cite{midas}, 
which allows us to regress a depth image from an RGB image in both indoor and outdoor environments without relying on an additional sensor modality and an adaptation stage.

To summarize, our problem and method are most relevant to two independent fields: single-modal cross-domain visual place classification and multi-modal sensor integration. However, the intersection of these two research fields has not been sufficiently explored yet. In the current work, this issue is explored by using a ``domain-invariant" monocular depth estimation as intermediate. To the best of our knowledge, no previous study has investigated in the above context.

\figF

\section{
Self-supervised Long-term Visual Place Classification 
}\label{sec:problem}

In 
long-term robot navigation, the training/retraining of a visual place classifier should be conducted in a completely self-supervised manner, without relying on external sensing devices such as GPS or 3D environment models. In this study, a 3-dof wheeled mobile robot is supposed, although this framework is sufficiently general to be extended to 6-dof vehicle applications such as drones. Nevertheless, the robot's workspace usually contains unmodeled three-dimensional undulations and elevation changes, such as small hills, which may affect visual recognition performance.

We focus on a simplified setup, single-session 
supervised training
and single-view classification. That is, it is assumed that a visual experience collected by 
a survey robot navigating the entire workspace in a single session is used as 
the sole supervision, and that 
the visual place classifier takes a single-view image as 
the sole query input. Nevertheless, this approach could be easily extended to multi-session supervision and multi-view self-localization setups, as in \cite{itsc2019fang}. 

\figB

The training stage starts with the robot navigating the target environment and collecting view sequences along the viewpoint trajectory in the training domain. It is assumed that 
the viewpoint trajectory has 
sufficiently long travel distance,
many loop closures,
which
allows 
sufficiently
accurate viewpoint reconstruction
via structure-from-motion, SLAM, and visual odometry.
Next, viewpoints are divided into place classes by spatially coarse partitioning of the robot workspace (Fig. \ref{fig:B}). 
Note that
the ground-truth viewpoint-to-class mapping 
is defined with respect to 
the training viewpoint trajectory
reconstructed,
without
assuming the availability of 
any GPS measurement.

We now formulate the classification task. Let $x_i$ be the view image at the $i$-th viewpoint, 
$y_i$ be the place class to which the viewpoint belongs, 
and the training data is expressed in the form $S^{train}$$=\{(x_i, y_i)\}$. Then, the training objective is to optimize the parameters of the classifier
\begin{equation}
y=f(x)
\end{equation}
using the training data $S^{train}$, so that the prediction performance for the test sample $x$$\in$$S^{test}$ in the unknown domain should be maximized.

The robot workspace is partitioned into a regular grid of 10$\times$10=100 place classes (Fig. \ref{fig:B}), for the following motives. (1) The grid-based place definition provides a flexible place definition for cross-domain self-localization scenarios. 
This is in contrast to 
the in-domain scenarios, 
such as the planet-scale place classification in \cite{planet}, 
where the spatial distribution of viewpoints is known in advance, 
allowing 
a more spatially efficient adaptive place partitioning. (2) The grid can be extended to unseen place classes found in new domains. For example, in \cite{icra2019sc}, an entropy-based discovery of unseen place classes is considered for a cross-domain place classification from an on-board Velodyne 3D scanner. (3) The grid-based place definition is often used for local/global path planning in visual robot navigation. For example, in \cite{cvmi2022kurauchi}, the place-specific knowledge of 
a visual place classifier is transferred to 
a reinforcement learning-based next-best-view 
path planner. (4) The number of place classes, a key hyperparameter, should be consistent with practical applications in the domain of NCLT dataset. The setting, 100, is consistent with the ``coarser'' grid cells in \cite{icra2019sc}, long-term knowledge distillation 
in \cite{itsc2019hiroki}, 
and active self-localization in \cite{cvmi2022kurauchi}.

We observe that compared to other image classification tasks such as object recognition, the visual place classification task has several unique and noteworthy properties:
(1) 
Viewpoint trajectory is not exactly the same between the training and deployment domains, even when the robot follows the same route. In fact, comparing the two extreme cases of navigating along the right and left edges of the route, the viewpoint positions are often more than 1 m apart. 
(2)
Differences in bearing often have a greater impact on prediction results than differences in place class, especially in typical 
outdoor workspaces where wide-open space scenes dominate and many objects are far from the 
robot body's turning center.
(3)
Due to differences in robot navigation tasks and changes in local traversability of the workspace, the routes in the training and test domains do not overlap completely. This 
yields unseen place classes,
which 
significantly complicates the problem.

\section{
Multi-modal Extension of Single-modal Classifier
}\label{sec:approach}

Our experimental setup,
multi-modal extension of single-modal classifier,
is specifically tailored for
the extension from RGB to RGB-D.
To this end, we consider a 
conventional setup of 
training 
a CNN as 
a single-modal RGB monocular image classifier
and use it as our baseline model, 
as in section \ref{sec:cnn}. 
It is known that such a monocular image classification task is significantly ill-posed due to the complex non-linear mapping of the 3D world to 2D images as well as domain shifts. To regularize the ill-posed problem, we introduce a monocular depth estimator as in section \ref{sec:midas} and further transform the depth image to a regularized HHA image as in section \ref{sec:hha}. Then, we train another single-modal HHA-image classifier CNN that takes the synthetic HHA images as input. Finally, the outputs of the two CNNs,
``RGB-Net" and ``HHA-Net",
are fused by an integration network with a multi-layer perceptron, which is then fine-tuned using the entire dataset as supervision, as detailed in section \ref{sec:mlp}.

\subsection{
Visual Classifier and Embedding
} \label{sec:cnn}

For the baseline classifier, 
we fine-tune a pretrained CNN, VGG16, 
to our datasets. VGG16 is a variant of CNN, proposed by the Visual Geometry Group of University of Oxford and the winner of the 2014 ILSVRC object identification algorithm \cite{vgg16}. 
It consists of 13 convolutional layers and 3 fully connected layers for a total of 16 layers. 
In this work, the CNN model 
\begin{equation}
y=f^{CNN}(x)
\end{equation}
is trained as a place classifier by fine-tuning the fully connected layers with the convolutional layers frozen.

In the proposed framework, the same CNN is also used as a means of image embedding:
\begin{equation}
f^{RGB}(x) 
= 
g^{RGB} \circ h^{RGB} (x), \label{eqn:rgb}
\end{equation}
where
$h^{CNN}$
is the embedding function.
It is well known that the 
fully-connected layer (FCL) signals of such a CNN can be viewed as an embedding of an input image to a discriminative feature vector. We performed a grid search with an independent validation set to find the best FCL
that most suits to our application. As a result, the second fully connected layer was found to be optimal. Therefore, it is decided to be used as the image embedding throughout all the experiments.

\subsection{
Monocular Depth Estimation
} \label{sec:midas}

We used MiDaS as a means of monocular depth estimation. MiDaS was originally presented by Ranftl et al \cite{midas}, 
to address the performance degradation of conventional monocular depth estimation models in cases where they were trained from insufficient datasets and therefore cannot generalize well to diverse environments, and to address the difficulty of large-scale capture of diverse depth datasets. In \cite{midas}, a strategy for combining complementary sources of data was introduced, and improved performance was demonstrated using a flexible loss function and a new strategy for principle-based data set mixing. Furthermore, the no-retraining property is obviously valuable for our cross-domain scenarios. Specifically, the MiDaS takes an RGB scene image $x$ as input and returns a pseudo depth image $y^{MiDaS}$:
\begin{equation}
y^{MiDaS}=f^{MiDaS}(x).
\end{equation}
Figure \ref{fig:F} shows examples of the estimated depth image.

\subsection{
Depth Image Encoding
} \label{sec:hha}

We further propose to encode the 1-channel depth image provided by the monocular depth estimation into a much more informative 3-channel HHA image. HHA is an image encoding method proposed by \cite{hha}, 
in order to represent each pixel of a given image by 3-channels, consisting of 
``Height above ground", 
``Horizontal disparity" and 
``Angle with gravity". 
The angle with the direction of gravity is estimated and then used to compute the height from the ground. The horizontal parallax for each pixel is obtained from the inversely proportional relationship with the original depth value. 

The overall algorithm is  an iterative process of updating the gravity vector. For the $t$-th iteration ($t\ge 1$):
\begin{enumerate}
\item
The input point cloud is split into a set $N_{||}$ of points parallel to the gravity vector and a set $N_{\perp}$ of points perpendicular to the gravity vector and the rest, where
\[
N_{||} = \{ n : \angle(n, g_{i-1}) < d 
\lor \angle(n, g_{i-1}) > \pi - d \}
\]
\[
N_{\perp} = \{ n :  \pi/2 - d < \angle(n, g_{i-1}) < \pi/2 + d \}
\]
The initial estimate for the gravity vector $g$ is the $y$-axis. 
For the variable $d$,
the setting of $d = \pi/4$ is used for $t\le 5$,
and 
the setting of $d=\pi/12$ is used for $t>5$.
\item
The gravity vector 
$g_i$
is updated by
\[
\min_{g:||g||_2=1} \sum_{n\in N_{\perp}} \cos^2 \left( \angle(n,g) \right) + \sum_{n\in N_{||}}\sin^2 \left(\angle(n,g)\right).
\]
\end{enumerate}

As a result, a given depth image is transformed into a 3-channel HHA image.
In an ablation study, we compared the original 1-channel depth image with the 3-channel HHA encoded image, in terms of the CNN classifier performance, and found that a large performance drop was found in the former case.

Given the HHA modality:
\begin{equation}
y^{DIE}=f^{DIE}(x),
\end{equation}
the same CNN and embedding architectures 
as (\ref{eqn:rgb})
are used for the HHA modality:
\begin{equation}
f^{HHA}(x) 
= 
g^{HHA} \circ h^{HHA} (x). 
\end{equation}

\subsection{
Multimodal Network
} \label{sec:mlp}

Two independent CNN models, called RGB-Net and HHA-Net, are trained respectively using the RGB image and HHA images as the input modalities, and then a pair of image embeddings from the CNN pair is integrated by an additional integration network. 
There are two roles we 
could
expect from this integration network.
One is a switching role, 
aiming at 
diagnosing 
inputs from RGB-Net and HHA-Net
to filter out invalid inputs.
This diagnostic problem is non-trivial. 
Note that this is because we only have two inputs, 
so even when 
we detect 
inconsistencies
between them, we cannot tell which one is 
invalid.
Another role is 
a weighted average
of inputs.
This mixing problem is easy to solve, at least naively. For example, a naive way would be to output equally weighted RGB-Net and HHA-Net. However, 
we observed that
this naive method was often useless,
and yielded worse performance than either RGB or HHA -Net.

Our proposal is to implement this mixing with a trainable multi-layer-perceptron (MLP). This strategy has often worked, as 
will be shown in the experimental section. Note that this use of MLP as a mixing function has also been successfully used in many contexts, such as multi-supervisor knowledge transfer \cite{gou2021knowledge}.
The MLP consists of three layers and each layer has 8192, 1024, and 100 neurons, respectively. 
The number of neurons for the input layer, 8192,
corresponds to the 
concatenation of 
the pair of 4096-dimensional embeddings from the two networks (i.e., 4096x2=8192).
The number of neurons for the output layer,
100,
corresponds to the number of place classes.

\subsection{Training}

Our framework employs several learnable parameters:
$f^{RGB}$, 
$f^{MiDaS}$, 
$f^{HHA}$, 
$f^{DIE}$, 
and $f^{MLP}$.
We assume
the parameters 
$f^{MiDaS}$
and
$f^{DIE}$
are domain invariant,
while
$f^{RGB}$, 
$f^{HHA}$, 
and
$f^{MLP}$
must be fine-tuned to the target domain.
Note that the model is trained efficiently by the following procedure.
\begin{enumerate}
\item
The CNN model $f^{RGB}$ is trained using the RGB images $X^{RGB}=S^{train}$ and the given ground-truth class labels.
\item
All the RGB images $X^{RGB}$ are transformed to HHA images $X^{HHA}$ by using the models $f^{MiDaS}$ and $f^{DIE}$.
\item
The CNN model $f^{HHA}$ is trained using the HHA images $X^{HHA}$ and the given ground-truth class labels.
\item
All the RGB images $X^{RGB}$ are fed to the trained 
embedding model $h^{RGB}$ to obtain embeddings $Y^{RGB}$.
\item
All the HHA images ${S}^{HHA}$ are fed to the trained 
embedding model $h^{HHA}$ to obtain embeddings $Y^{HHA}$.
\item
All the corresponding pairs from $Y^{RGB}$ and $Y^{HHA}$ are 
concatenated
to obtain a training set $Y^{MLP}$ for MLP.
\item
MLP is finally trained using the set $Y^{MLP}$ as supervision.
\end{enumerate}

\figC

\figD

\figE

\section{Experiments}\label{sec:exp}

\subsection{Dataset}

The NCLT, one of the most popular datasets for cross-season visual robot navigation, was used for performance evaluation. The NCLT dataset is a collection of outdoor images collected 
by a Segway vehicle
every other week from January 8, 2012 to April 5, 2013 at the University of Michigan North Campus. For each dataset, the robot travels indoor and outdoor routes on the university campus, while encountering various types of static and dynamic objects, such as 
desks, chairs, pedestrians and bicycles, and also experiences long-term cross-dataset changes such as snow cover, weather changes, and building renovations. In this work, the on-board front-facing camera of the vehicle was used as the main modality. Also, the associated GPS data was used as the ground-truth for the self-localization task. Figure \ref{fig:B} 
shows the bird's eye view of the robot workspace and viewpoint trajectories.

Four datasets with IDs, 2012/03/31 (SP), 2012/08/04 (SU), 2012/11 /17 (AU), and 2012/01/22 (WI) are used for the current experiments. The number of images in these $N=4$ datasets is 26,364, 24,138, 26,923 and 26,208, respectively. The image was resized from the original size of 1,232$\times$1,616 pixels to 256$\times$256 pixels. Example images in each dataset are shown in Figure \ref{fig:C}. Different experiments were conducted by using each of all the $N(N-1)=12$ pairings of the four datasets as the training-test dataset pair.

The robot workspace is defined by the bounding box of all viewpoints and partitioned into a 10 $\times$ 10 grid of place classes, before the training and test stages.

\subsection{Results}

As mentioned in Section \ref{sec:cnn}, 
VGG16 was used as a 
comparative method. This Vgg16 model is 
exactly the same 
as the Vgg16 that the proposed method uses as a feature extractor, with the exactly same training procedure, conditions and hyperparameters.

The performance is evalauted in terms of top-1 accuracy, which is defined as the ratio of successful test samples over the entire test set. Here, a test sample is judged as successful if and only if its maximum likelihood class is consistent with the ground-truth class.

For an ablation study, we also trained an alternative baseline single-modal CNN model, ``HHA-Net",
which 
uses the HHA-images instead of the RGB-images as the CNN input, in the same procedure as the aforementioned baseline model.

Table \ref{tab:A}
shows the performance results. One can see that the proposed method outperforms the comparative methods, RGB-Net and HHA-Net, in all the 12 combinations of training and test datasets and recognition performance improves by from 3.9pt to 13.5pt. 

Several examples of the input image, the ground-truth class image, and the predicted class image for successful and failure examples respectively are shown in Figs. \ref{fig:D} and \ref{fig:E}.
In both figures, the columns are, from left to right, the (RGB, HHA) image pair of the test sample, the place class that received the highest likelihood visualized by a training sample image pair, and the ground -truth image. It can be observed that the proposed method intelligently identifies the shapes of mountains and roads, the presence or absence of buildings, etc., and uses them for classification. On the other hand, classification often fails in confusing scenes where even a human could get lost. We also encountered errors in mistaking buildings for trees, which could be compensated for by introducing semantic features in future studies.

It could be concluded that the proposed method, multimodal formulation of single-modal 
visual place classification,
led to significant improvements in performance and robustness.

\tabA

\section{
Concluding Remarks
}\label{sec:conclusions}

In this work, we revisited the challenging problem of cross-domain visual place classification from a new perspective of multimodal RGB-D fusion. The experimental setup was based on two domain-invariant schemes. One is the pseudo-multimodal fusion scheme that is expected to inherit the domain invariance ability of multi-modal senser integration approach, without requiring additional sensing device. The other is the introduction of domain-invariant pseudo-depth measurement called domain-invariant monocular depth estimation. A realistic framework for information processing and information fusion of these multimodal data was presented and validated in a 
practical long-term 
robot navigation scenario. 
It was confirmed 
that the proposed method clearly contributes to the performance improvement in all the datasets considered here.

\bibliographystyle{IEEEtran} 
\bibliography{reference}

\end{document}